# RAD: Retrieval-Augmented Decision-Making of Meta-Actions with Vision-Language Models in Autonomous Driving

Yujin Wang[1]     Quanfeng Liu[1]     Zhengxin Jiang[1]     Tianyi Wang[2]
Junfeng Jiao[3]     Hongqing Chu[1,*]     Bingzhao Gao[1]     Hong Chen[1]
[1]*Tongji University*     [2]*Yale University*     [3]*University of Texas at Austin*
chuhongqing@tongji.edu.cn

## Abstract

*Accurately understanding and deciding high-level meta-actions is essential for ensuring reliable and safe autonomous driving systems. While vision-language models (VLMs) have shown significant potential in various autonomous driving tasks, they often suffer from limitations such as inadequate spatial perception and hallucination, reducing their effectiveness in complex autonomous driving scenarios. To address these challenges, we propose a retrieval-augmented decision-making (RAD) framework, a novel architecture designed to enhance VLMs' capabilities to reliably generate meta-actions in autonomous driving scenes. RAD leverages a retrieval-augmented generation (RAG) pipeline to dynamically improve decision accuracy through a three-stage process consisting of the embedding flow, retrieving flow, and generating flow. Additionally, we fine-tune VLMs on a specifically curated dataset derived from the NuScenes dataset to enhance their spatial perception and bird's-eye view image comprehension capabilities. Extensive experimental evaluations on the curated NuScenes-based dataset demonstrate that RAD outperforms baseline methods across key evaluation metrics, including match accuracy, and F1 score, and self-defined overall score, highlighting its effectiveness in improving meta-action decision-making for autonomous driving tasks.*

## 1. Introduction

In recent years, the race towards fully autonomous vehicles has spurred extensive research into robust decision-making approaches, a fundamental task in autonomous driving systems [26, 41, 49]. Ensuring safe and efficient motion planning requires continuous interpretation of dynamic environments, real-time reasoning under uncertainty, and efficient integration of vast amounts of multimodal data [28].

Traditional autonomous driving systems adopt a modular development strategy, in which perception, prediction, planning, and control are developed and optimized independently before being integrated into the vehicle system [15, 47]. However, as the information flow propagates across these modules, errors and delays can accumulate, potentially leading to suboptimal or even unreasonable driving decisions. To further mitigate these errors and improve computational efficiency, end-to-end autonomous driving has emerged as a prominent research direction [7, 8].

End-to-end refers to a model that directly receives input from sensor data (e.g., cameras, LiDAR) and directly outputs vehicle planning decisions. In recent studies [11, 18, 22], end-to-end autonomous driving algorithms have demonstrated their superiority in both simulation environments and real-world road tests. Moreover, the emergence of foundation models provides a promising solution to enhance motion planning performance, improve generalization across diverse scenarios, and increase interpretability in end-to-end autonomous driving [13, 16, 29, 38]. Trained on huge amounts of human knowledge, these models exhibit advanced comprehension and reasoning capabilities, highlighting the immense potential of artificial intelligence in complex decision-making tasks. Integrating such foundation models into autonomous driving systems could facilitate the development of human-like driving behaviors, advancing the field toward safer and more adaptable autonomous vehicles.

Autonomous driving tasks require models with robust visual perception capabilities, making vision-language models (VLMs) particularly well-suited for this domain. VLMs trained on large-scale data often demonstrate strong reasoning capabilities, enabling them to infer the evolution of complex driving scenarios. Current research [19, 31, 33, 34, 36] has focused on fine-tuning pre-trained VLMs using visual question-answer (VQA) pairs composed of scene images and corresponding driving actions. This approach enables VLMs to generate feasible trajectories, enhancing their applicability in real-world autonomous driving tasks.

However, fine-tuning or even full-scale fine-tuning of VLMs using large-scale datasets requires substantial com-putational resources. Additionally, deploying VLMs with an extremely large number of parameters on vehicle-end hardware poses significant constraints. To address these challenges, retrieval-augmented generation (RAG) has emerged as a promising approach to enhance the decision-making capabilities of VLMs by incorporating external knowledge bases [14, 42]. The core idea of RAG is to augment generative models with a retrieval module that dynamically retrieves relevant textual information during the generation process. In vision-language tasks, RAG can effectively mitigate limitations caused by knowledge scarcity. By integrating external knowledge bases, models can not only extract information from images but also retrieve supplementary knowledge, thereby improving the robustness and accuracy of the generated outputs. Although the direct application of RAG to the decision-making process in autonomous driving remains limited, an increasing number of studies have explored its potential in specific tasks such as scene understanding and regulation retrieval [4, 20, 46].

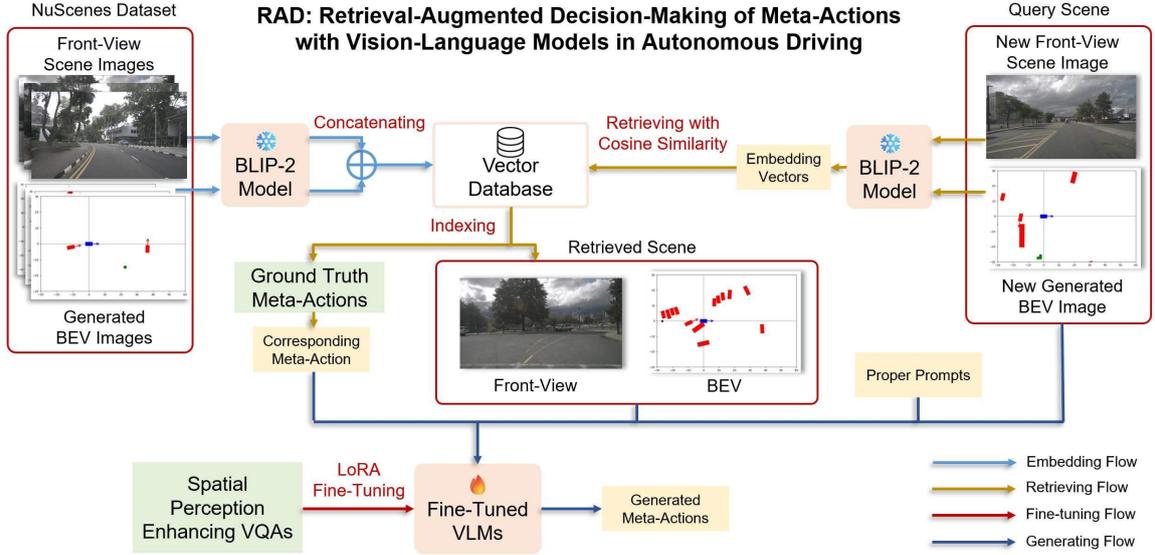

Figure 1. The overview of our RAD method. The framework consists of four working flows, namely embedding flow, retrieving flow, fine-tuning flow and generating flow. The embedding flow encodes front-view images and BEV images into a vector database. Given a query scene, the retrieving flow retrieves the most similar scene from the database. The fine-tuning flow involves fine-tuning VLMs to enhance spatial perception and BEV image comprehension. The generating flow guides VLMs in generating contextually appropriate meta-actions according to the query scene, the retrieved scene, its ground truth meta-action, and proper prompts.

In this work, we propose a retrieval-augmented decision-making (RAD) framework, introducing a novel approach to assist VLMs in generating meta-actions using RAG for the first time, as depicted in Figure 1. The main research contributions of this work are outlined as follows:

- **Pre-Training VLMs for Spatial Perception Tasks:** We construct obstacle perception tasks based on the NuScenes dataset [3], incorporating VQA pairs designed to capture obstacle categories, positions, and other spatial information. This pre-training process enables VLMs to explicitly learn key geometric features such as the locations and sizes of obstacles, leading to improved performance in spatial perception tasks.
- **Establishing an External Knowledge Base with NuScenes Ground Truth Data:** We select a subset of scenes containing navigation information, historical trajectory data, and future meta-action ground truth. Furthermore, we generate bird's-eye view (BEV) images corresponding to the scene images. The surround-view images from these scenes are then encoded into vector representations using BLIP-2 [25], alongside the BEV images, to form the knowledge base.
- **Developing a Retrieval and Generation Pipeline for Meta-Action Decision-Making using Fine-Tuned VLMs and RAG:** We employ cosine similarity to retrieve the most similar scene from the external knowledge base including the front-view image of the current scene. The corresponding six surround-view images, speed information, navigation data, and ground truth trajectory are then used as auxiliary inputs, guiding the VLM in generating a trustworthy planning trajectory for the current scene.

The remainder of this paper is organized as follows: In Section 2, the detailed literature review is conducted. In Section 3, four working flows of the proposed RAD framework are introduced. In Section 4, comparative experiments and ablation studies are designed. Section 5 summarizes the work and discusses future research directions.

## 2. Related Works

### 2.1. Multimodal Large Language Models in Autonomous Driving

Utilizing multimodal large language models (MLLMs) in autonomous driving enhances decision-making by leveraging their extensive knowledge and reasoning capabilities through multisource information such as vision, language, and rules, significantly improving scene understanding, strategy generation, and interpretability. DriveMLM [39] employed MLLMs to generate high-level behavioral decisions (e.g., lane-changing, deceleration, acceleration, etc.), which were then integrated with traditional motion planning modules, balancing flexibility and interpretability. "Drive as you speak" [10] enriched large language models (LLMs) with comprehensive environmental data from different vehicle modules, leading to safer decisions. "Driving with LLMs" [6] introduced a LLM that generated 10,000 driving scenarios for agent training. "Drive like a human" [13] demonstrated LLMs' capabilities of understanding and interacting with environments in closed-loop systems, effectively navigating long-tail autonomous driving scenarios. DriveVLM [37] adopted a multistage reasoning chain that combined scene description, dynamic analysis, and hierarchical planning. Additionally, DriveVLM-Dual incorporated traditional 3D perception algorithms to ensure both cognitive depth and real-time control. Pix2Planning [30] formulated planning as an autoregressive sequence prediction problem, using a vision-language Transformer to generate trajectory points. VLP [31] incorporated linguistic descriptions into the training process and aligned them with visual features, significantly improving cross-city and cross-scenario generalization. To enhance interpretability, some studies [44, 45] introduced "future trajectory images", which were processed by multimodal models to generate natural language explanations. Senna [23] further refined the decision-making process by separating high-level meta-actions from low-level trajectory predictions. In this framework, VLMs first produced directional or speed-level decisions before end-to-end models executed precise paths, thereby achieving a hierarchical strategy that was similar to human driving behaviors.

However, these methods are prone to hallucination, a limitation arising from the reliance of MLLMs on learned associations between visual inputs and language-based reasoning. As a result, they may misinterpret ambiguous or occluded objects, leading to incorrect high-level decision-making. This issue becomes particularly critical in long-tail scenarios, where the model encounters rare or underrepresented driving conditions not well-covered in the training data. Such misinterpretations can ultimately compromise the reliability and safety of the autonomous driving system.

### 2.2. Retrieval-Augmented Generation in Vision-Language Models

In vision-language tasks, RAG mitigates knowledge limitations by leveraging external knowledge bases, enabling models to extract insights from images while supplementing them with retrieved contextual data. This dual approach significantly helps mitigate model hallucination and improve planning accuracy. Jiang et al. [24] introduced a RAG-based framework for VLMs, demonstrating its effectiveness in complex tasks requiring extensive background knowledge. Their study underscored the limitations of conventional end-to-end VLMs when faced with knowledge deficiencies, whereas RAG facilitated richer contextual integration, enhancing both reasoning and generation. Building on this, Shao et al. [35] further investigated RAG's role in VQA tasks, showing that combining retrieval mechanisms with pre-trained VLMs significantly strengthened model performance in complex reasoning scenarios. Additionally, Ram et al. [32] examined RAG's impact on pre-training and fine-tuning, illustrating that incorporating large-scale external data sources during pre-training improved downstream performance by enhancing cross-modal reasoning, particularly in retrieval-based tasks. Meanwhile, Zheng et al. [50] emphasized RAG's broader advantages, particularly in improving generative flexibility and adaptability in multimodal tasks. Their findings highlighted RAG's effectiveness in handling scenarios lacking sufficient annotations or domain-specific knowledge, reinforcing its potential in bridging knowledge gaps for more informed and context-aware model outputs. Hussien et al. [20] illustrated how RAG-augmented VLMs enhanced cross-modal retrieval, particularly by strengthening associations between images and textual data. For performance optimization, Yuan et al. [46] introduced a dynamic knowledge retrieval mechanism, emphasizing real-time adjustments in retrieval and generation processes based on task-specific requirements. This adaptive approach allowed RAG to selectively retrieve the most relevant background knowledge, improving performance across various multimodal applications. Cai et al. [4] developed a traffic regulation retrieval agent based on RAG, enabling automatic retrieval of relevant traffic rules and guidelines based on the ego vehicle's status. Moreover, Cui et al. [12] incorporated a RAG-based memory module that continuously learned takeover preferences through human feedback to enhance motion planning.

Despite its strong potential, research on directly utilizing RAG to guide VLMs in meta-action decision-making remains limited. To address this gap, we propose the RAD framework, which, for the first time, integrates RAG with pre-training for spatial perception capabilities, enabling more effective decision-making of meta-actions.

## 3. Methodology

As shown in Figure 1, the proposed RAD framework comprises four work flows: embedding flow, retrieving flow, fine-tuning flow and generating flow. Among these, the fine-tuning flow operates independently, as its primary objective is to enhance the spatial perception capabilities of VLMs through separate fine-tuning. In the embedding flow, BEV images are generated to correspond with front-view scene images from the NuScenes dataset. These image pairs are encoded into a vector space using a frozen BLIP-2 model and the separate embeddings are then concatenated and stored in a vector database. In the retrieving flow, a new front-view image and its corresponding BEV image serve as a query. These images are encoded into the vector space using the same frozen BLIP-2 model. Cosine similarity is then computed between the query images and those stored in the database, enabling the retrieval of the most similar scene from the database. Furthermore, based on the relative positional relationships between consecutive scenes in the NuScenes dataset, the ground truth meta-actions executed in each scene can be extracted. Finally, in the generating flow, the query scene, retrieved scene, its ground truth meta-action, and proper prompts serve as inputs to the VLMs. These inputs guide the model to make decisions and generate meta-actions, ensuring more accurate and context-aware autonomous driving behaviors.

All the extracted meta-actions are shown as follows:

(1) Speed up (rapidly)       (2) Slow down (rapidly)
(3) Turn left/right          (4) Drive along the curve
(5) Turn around              (6) Change lane to the left/right
(7) Reverse                  (8) Shift slightly to the left/right
(9) Stop                     (10) Go straight constantly/slowly

### 3.1. The Fine-Tuning Flow

Making precise meta-action decisions in autonomous driving requires an accurate understanding of the environment. If a model lacks sufficient spatial perception capabilities, it may fail to construct a reliable environmental representation, potentially leading to obstacle avoidance failures in meta-action decision-making. VLMs typically rely on monocular or surround-view camera inputs and estimate depth information from single-frame images. However, in long-trail scenarios, monocular vision exhibits significant depth estimation errors [5]. Experimental results on the NuScenes dataset indicate that existing VLMs generally lack robust spatial perception, which severely impacts the safety of decision-making and motion control [43].

To address the aforementioned challenges, VLMs should first undergo fine-tuning to enhance their spatial perception capabilities. The structure of VLMs typically consists of a vision encoder and an LLM. In this work, we focus on fine-tuning only the LLM component to enhance its spatial perception. We utilize the NuScenes dataset to generate a specified dataset for spatial perception enhancement, following the process illustrated in Figure 2.

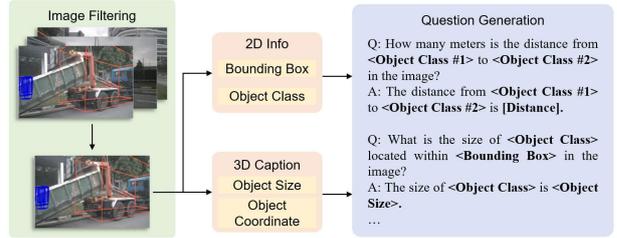

Figure 2. The process of generating a dataset for spatial perception enhancement based on the NuScenes dataset

During the image filtering process, it is necessary to ensure the uniqueness of the VQA pairs by cross-referencing the annotated data from the origin NuScenes dataset. The generated dataset for fine-tuning includes over 100,000 training samples, covering key spatial perception tasks such as object class recognition, object distance estimation and object size estimation.

For spatial perception enhancement fine-tuning, the loss function for a single sample is defined as follows:

$$J = \frac{1}{N} \sum_{i=1}^{N} \left[ \lambda_{1,i} \left( -\sum_{c=1}^{n} y_{c,i} \log(p_{c,i}) \right) \right.$$
$$\left. + \lambda_{2,i} \left( -\frac{1}{3} \sum_{j=1}^{3} (z_{j,i} - z_{j,i}^{*})^2 \right) + \lambda_{3,i} (x_i - x_i^{*})^2 \right] \quad (1)$$

where, $N$ is the batch size during fine-tuning; $\lambda_{1,i}$ is the loss identifier for object class recognition in the $i$-th sample (if there is a corresponding class, $\lambda_{1,i}$ will be set to 1; and otherwise, $\lambda_{1,i}$ will be set to 0); $\lambda_{2,i}$ is the loss identifier for object size estimation; $\lambda_{3,i}$ is the loss identifier for object distance estimation; $n$ is the total number of classes in the classification task; $y_{c,i}$ is the label for the $i$-th sample belonging to class $c$, represented by one-hot encoding; $p_{c,i}$ is the probability of the $i$-th sample being classified as class $c$ by the model; $z_{j,i}$ is the output size of the $i$-th sample in the $j$-th dimension from the model; $z_{j,i}^{*}$ is the ground truth size of the $i$-th sample in the $j$-th dimension; $x_i$ is the model's output for the distance from the $i$-th sample to the reference object; and $x_i^{*}$ is the ground truth distance from the $i$-th sample to the reference object.

In this work, we fine-tune a series of VLMs, primarily from the Qwen family [1, 2, 9], using low-rank adaption (LoRA) [17, 51]. The overall training is conducted for

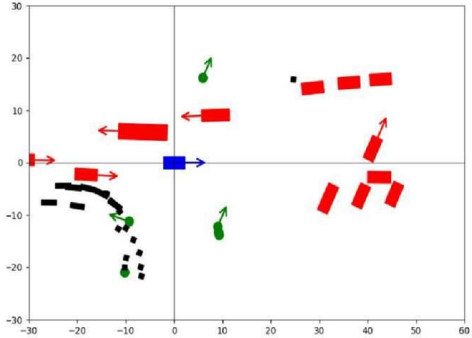

**System:** This image illustrates the BEV view of a driving scene, showing the area of 60 meters ahead, 30 meters behind, 30 meters left and right of the ego vehicle. The units of longitudinal and lateral coordinates are meters. The ego vehicle is located at the center [0,0], represented by a blue rectangle. The red rectangles represent the objects of vehicle type, including cars, trucks, etc. If there is an arrow on the red rectangle, it means that it will move in the direction of the arrow. The green dots represent pedestrians, and the green arrows also indicate the moving direction. Black dots are static obstacles, including roadblocks, traffic lights, etc.

**Question 1:** What kind of object (pedestrian, vehicle, or static obstacle) is located within the coordinate [7.6,8.9] in this image?
**Answer 1:** There is a vehicle located within the coordinate [7.6,8.9].
**Question 2:** What is the central position coordinate of the left-front static obstacle in this image? The result retains one decimal place after the decimal point.
**Answer 2:** The central position coordinate of the left-front static obstacle is [24.5,17.2].
**Question 3:** What is the distance from the left-front static obstacle to the left pedestrian in this image? The result retains one decimal place after the decimal point.
**Answer 3:** The distance from the left-front static obstacle to the left pedestrian is 16.3 m.

Figure 3. The fine-tuning VQA paradigm for BEV image understanding

three epochs. Additionally, following the BEVFormer [27], we generate BEV images from the existing surround-view images in the NuScenes dataset. Intuitively, incorporating BEV images helps the model better understand the relative spatial relationships of objects in driving scenes. Therefore, it is also necessary to train VLMs to recognize and interpret BEV images effectively. The fine-tuning paradigm, as illustrated in Figure 3, follows a similar approach to the VQA pair construction method based on ground truth information to develop a robust ability to understand BEV images.

### 3.2. The Embedding Flow

In the embedding flow, we encode front-view images from the NuScenes dataset along with the pre-generated BEV images into a unified vector space. Since this embedding operation does not involve cross-modal content, the frozen BLIP-2 model weights can be directly utilized, ensuring computational efficiency and consistency. To maintain the one-to-one correspondence between front-view images and BEV images, their embedding vectors are concatenated within this flow. The resulting concatenated vectors are then uniformly stored in an indexed vector database.

### 3.3. The Retrieving Flow

The core of the retrieving flow lies in the computation of cosine similarity. Given two image embeddings $\mathbf{v}_i$ and $\mathbf{v}_j$, cosine similarity is defined as:

$$similarity_{i,j} = \frac{\mathbf{v}_i \cdot \mathbf{v}_j}{\|\mathbf{v}_i\|\|\mathbf{v}_j\|} \quad (2)$$

where, $\| * \|$ represents the Euclidean norm.

The main framework of the retrieving flow is illustrated in Figure 4. For a new scene, we first generate its BEV images from the surround-view images. The front-view image and BEV image of the new scene jointly trigger a query scene. The embeddings for the new front-view image and BEV image are then extracted using the frozen BLIP-2 model. Since the vector database stores concatenated embedding vectors, the embeddings for the front-view image and BEV image are retrieved through length decomposition. The cosine similarity between the new front-view image embeddings and those stored in the database is computed and denoted as $similarity_{fv}$. Similarly, the cosine similarity between the new BEV image embeddings and those stored in the database is computed and denoted as $similarity_{bev}$. To flexibly adjust the retrieval preference toward either the front-view image or the BEV image, a hyperparameter $\omega$ is introduced. In this work, $\omega$ is set to 0.5 as a balanced weight for retrieval. The overall similarity could be calculated as follows:

$$similarity = (1 - \omega) \cdot similarity_{fv} + \omega \cdot similarity_{bev} \quad (3)$$

The scene with the highest overall similarity is then retrieved from the vector database. Using its index, we can obtain the corresponding front-view image, BEV image, and pre-extracted ground truth meta-action.

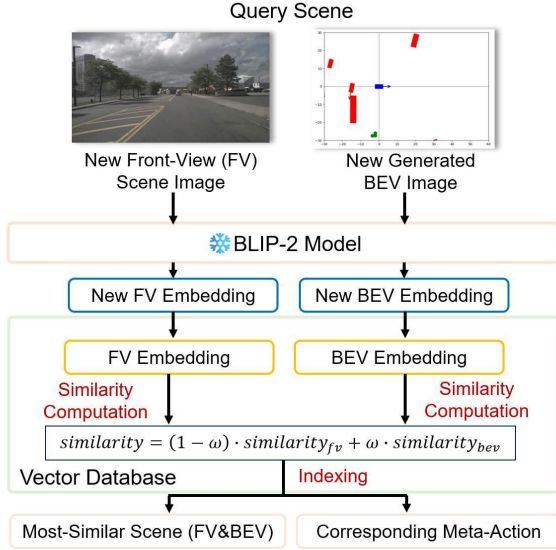

Figure 4. The main framework of the retrieving flow

### 3.4. The Generating Flow

In the generating flow, we primarily employ prompt engineering to guide VLMs in reasoning based on the retrieved scene and its corresponding meta-action, enabling them to make accurate meta-action decisions for the new scene. The prompts should be divided into two key components:

- **System Prompt:** Guide VLMs to make meta-action decisions based on the provided images.
- **RAG-Specific Prompt:** Instruct VLMs to understand the retrieved scene images and corresponding meta-actions.

For this process, we primarily use the Qwen series of VLMs, as they support multiple image inputs, making prompt design more flexible and effective. With structured and well-designed prompts, the VLMs analyze the front-view image and BEV image of the current scene, ultimately generating a single meta-action as the final output.

## 4. Experiments

### 4.1. Dataset Preparation

We divide the 34,000 scenes from the NuScenes dataset into three subsets: 10,000 scenes are allocated for fine-tuning VLMs, focusing on enhancing spatial perception and BEV image understanding; 20,000 scenes are embedded in the vector database as prior information; and the remaining 4,000 scenes serve as the test set, used to evaluate the framework's effectiveness and the model's overall performance.

### 4.2. Evaluation Metrics

To assess the performance, we employ traditional classification metrics such as accuracy, precision, recall and F1 score. Additionally, we introduce a customized partial match score to account for semantically similar but not entirely identical cases. Finally, we utilize a weighted method to compute a comprehensive performance score.

We firstly adopt $ExactMatchAccuracy$ to evaluate whether the model provides a fully correct meta-action for a given scene, which is formally defined as follows:

$$ExactMatchAccuracy = \frac{N_{match}}{N_{total}} \quad (4)$$

where, $N_{match}$ is the number of scenes where the generated meta-actions exactly match the ground truth; and $N_{total}$ is the total number of scenes.

For each meta-action, $Precision$, $Recall$, and $F1$ can be used as evaluation metrics, which are defined as follows:

$$Precision_i = \frac{TP_i}{TP_i + FP_i} \quad (5)$$

$$Recall_i = \frac{TP_i}{TP_i + FN_i} \quad (6)$$

$$F1_i = \frac{2 \times Precision_i \times Recall_i}{Precision_i + Recall_i} \quad (7)$$

where, $TP_i$ is the true positives, and the number of scenes where the generated meta-actions are $i$ and the ground truth are also $i$; $FP_i$ is the false positives, and the number of scenes where the generated meta-actions are $i$ but the ground truth are not $i$; $FN_i$ is the false negatives, and the number of scenes where the generated meta-actions are not $i$ but the ground truth are $i$.

To evaluate the overall performance across different meta-actions in the test set, $Macro - F1$ and $Weighted - F1$ scores are introduced. $Macro - F1$ is the unweighted average of $F1$ scores across all meta-actions, while $Weighted - F1$ is the weighted average of $F1$ scores, which are defined as:

$$Macro - F1 = \frac{1}{K} \sum_{i=1}^{K} F1_i \quad (8)$$

$$Weighted - F1 = \frac{1}{N_{total}} \sum_{i=1}^{K} n_i F1_i \quad (9)$$

where, $K$ represents the total number of meta-actions, which is set to 15; and $n_i$ represents the number of scenes where the ground truth meta-action is $i$.

To account for the semantic similarity between certain meta-actions, we introduce a $PartialMatchScore$. Specifically, meta-actions involving leftward maneuvers—such as *turn left*, *change lane to the left* and *shift*

Table 1. Comparison among different baselines and our RAD method

| Method | Exact Match Accuracy | Macro-F1 | Weighted-F1 | Partial Match Score | Overall Score |
|---|---|---|---|---|---|
| Lynx (Fine-tuning)[48] | 0.1524 | 0.0167 | 0.0653 | 0.2768 | 0.1327 |
| CogVLM (Fine-tuning)[40] | 0.2178 | 0.0204 | 0.1105 | 0.3563 | 0.1846 |
| DriveLM (on LLaMA-LoRA-BIAS-7B)[36] | 0.1455 | 0.0448 | 0.1203 | 0.3028 | 0.1518 |
| DriveLM (on LLaMA-BIAS-7B)[36] | 0.1896 | 0.0409 | 0.1212 | 0.3425 | 0.1693 |
| DriveLM (on LLaMA-CAPTION-7B)[36] | 0.2034 | 0.0380 | 0.1080 | 0.3952 | 0.1896 |
| GPT-4o (Official API)[21] | 0.2994 | 0.1127 | 0.2288 | 0.4377 | 0.2756 |
| DriveVLM[37] | 0.3743 | 0.1671 | 0.3325 | 0.5462 | 0.3589 |
| DriveVLM-Dual (cooperating with VAD[22])[37] | 0.4016 | 0.1854 | 0.3506 | 0.5613 | 0.3801 |
| **RAD (Ours, on Qwen-VL-2.5-7B)** | **0.4096** | **0.1907** | **0.3813** | **0.5870** | **0.3956** |

*slightly to the left*—are classified under the **left group**, while analogous rightward actions form the **right group**. Similarly, meta-actions indicating forward motion at varying speeds are categorized accordingly, with *go straight slowly*, *slow down*, and *slow down rapidly* mapping to the **deceleration group**, while both *speed up* and *speed up rapidly* mapping to **acceleration group**. Furthermore, unique behaviors such as *go straight constantly*, *turn around*, *reverse*, *stop*, and *drive along the curve* are collectively assigned to a separate **unique group**. If the generated meta-actions and the ground truth meta-actions are not identical but belong to the same semantic group (excluding the **unique group**), they are considered partially matched. Thus, the semantic similarity $S$ is defined as follows:

$$S(i, \hat{i}) = \begin{cases} 1, & \text{if } \hat{i} \text{ is the same as } i. \\ 0.5, & \text{if } \hat{i} \text{ partially matches } i. \\ 0, & \text{if } \hat{i} \text{ totally differs from } i. \end{cases} \quad (10)$$

where, $i$ is the ground truth meta-action in one scene; and $\hat{i}$ is the generated meta-action.

Then, the average $PartialMatchScore$ is obtained by averaging across all scenes:

$$PartialMatchScore = \frac{1}{N_{total}} \sum_{k=1}^{N_{total}} S(i_k, \hat{i_k}) \quad (11)$$

Finally, different weights are assigned to each metric to derive the comprehensive scoring formula $OverallScore$:

$$\begin{aligned} OverallScore = &\; \alpha \cdot ExactMatchAccuracy \\ &+ \beta \cdot Macro-F1 \\ &+ \gamma \cdot Weighted-F1 \\ &+ \delta \cdot PartialMatchScore \end{aligned} \quad (12)$$

where, $\alpha$ is set to 0.4; $\beta$, $\gamma$, and $\delta$ are all set to 0.2, which could be adjusted according to specific tasks.

### 4.3. Comparative Experiments

We evaluate the performance of our proposed RAD framework on Qwen-VL-2.5-7B VLM and compare it against several other state-to-the-art baseline methods: Lynx [48], CogVLM [40], DriveLM [36], GPT-4o [21] and DriveVLM [37]. Table 1 presents a thorough quantitative comparison between our proposed RAD and these baselines across multiple evaluation criteria. Our RAD consistently outperforms all baseline methods, demonstrating clear advantages in meta-action decision-making for autonomous driving. In particular, RAD achieves an $ExactMatchAccuracy$ of 0.4096, substantially outperforming DriveVLM-Dual's 0.4016, and attains an $OverallScore$ of 0.3956 compared to DriveVLM-Dual's 0.3801.

A deeper analysis of the remaining metrics further underscores RAD's strengths. $Macro-F1$, a balanced measure of model performance across all classes, achieves 0.1907, well above DriveVLM-Dual's 01854. Meanwhile, $Weighted-F1$ of 0.3813 indicates its effectiveness in scenarios where class imbalances exist, significantly outperforming all baselines and reflecting RAD's notable capabilities to handle diverse datasets. Also, $PartialMatchScore$ of 0.5870 also highlights RAD's fine-grained generative capability, which suggests that RAD not only excels at producing entirely correct answers, but also consistently captures partially correct information, an essential trait for more nuanced or multi-faceted decision-making tasks.

The poor performance of the baseline methods is mainly due to their lack of task-specific training. As a result, these models exhibit limited spatial perception capabilities and poor BEV image comprehension. Additionally, the parameter size constraints and version limitations of the base models used in these baselines hinder their ability to achieve optimal results. However, RAD's superior performance over GPT-4o across all metrics demonstrates the feasibility of specialized VLMs with smaller parameter sizes that rival or even surpass large-scale general-purpose models in complex and domain-specific tasks.

In summary, the results in Table 1 validate the efficacy and robustness of our RAD model. Through a combination of architectural innovations and targeted training strategies, RAD not only achieves profound performance across multiple metrics but also provides insights into how specialized

Table 2. Ablation studies on fine-tuning VLMs and RAG pipeline

| VLMs | Method | Exact Match Accuracy | Macro-F1 | Weighted-F1 | Partial Matching Score | Overall Score |
|---|---|---|---|---|---|---|
| Qwen-VL-2-2B[9] | Vanilla | 0.2188 | 0.0358 | 0.1013 | 0.4353 | 0.2020 |
| | Vanilla + RAG | 0.2145 | 0.1049 | 0.2278 | 0.4319 | 0.2387 |
| | Fine-tuning | 0.1543 | 0.0528 | 0.1194 | 0.3017 | 0.1565 |
| | **Fine-tuning + RAG** | **0.2610** | **0.1302** | **0.2556** | **0.4538** | **0.2723** |
| Qwen-VL-2-7B[9] | Vanilla | 0.2866 | 0.0654 | 0.1721 | 0.4941 | 0.2609 |
| | **Vanilla + RAG** | 0.3404 | **0.1460** | **0.3235** | **0.5424** | **0.3385** |
| | Fine-tuning | 0.2908 | 0.0717 | 0.1986 | 0.4562 | 0.2616 |
| | Fine-tuning + RAG | **0.3446** | 0.1460 | 0.3011 | 0.5213 | 0.3315 |
| Qwen-VL-2.5-3B[2] | Vanilla | 0.1318 | 0.0366 | 0.0955 | 0.3886 | 0.1568 |
| | Vanilla + RAG | 0.1240 | 0.0298 | 0.0814 | 0.3866 | 0.1491 |
| | Fine-tuning | 0.2164 | 0.0531 | 0.1398 | 0.3949 | 0.2041 |
| | **Fine-tuning + RAG** | **0.2539** | **0.1075** | **0.2090** | **0.4520** | **0.2552** |
| **Qwen-VL-2.5-7B**[2] | Vanilla | 0.2849 | 0.0644 | 0.1715 | 0.4893 | 0.2590 |
| | Vanilla + RAG | 0.3581 | 0.1981 | 0.3386 | 0.5544 | 0.3615 |
| | Fine-tuning | 0.3482 | 0.1085 | 0.2885 | 0.5360 | 0.3259 |
| | **Fine-tuning + RAG** | **0.4096** | **0.1907** | **0.3813** | **0.5870** | **0.3956** |

VLMs can excel in intricate autonomous driving tasks.

### 4.4. Ablation Studies

In our ablation studies, we mainly investigate the impacts of fine-tuning VLMs and RAG pipeline for spatial perception enhancement based on Qwen-VL-2-2B [9], Qwen-VL-2-7B [9], Qwen-VL-2.5-3B [2] and Qwen-VL-2.5-7B [2] models. The performance of VLMs is evaluated using four distinct methods: vanilla (no fine-tuning), vanilla combined with RAG, only fine-tuning, and fine-tuning combined with RAG (our proposed RAD method).

The results presented in Table 2 indicate that the combination of fine-tuning and RAG consistently achieves the highest scores across all evaluation metrics, including $ExactMatchAccuracy$, $Macro-F1$, $Weighted-F1$, $PartialMatchScore$, and $OverallScore$, for all model variants. Specifically, for Qwen-VL-2.5-7B, our RAD method achieves the highest $OverallScore$ of 0.3956, marking a significant improvement over methods that deploy either fine-tuning or RAG separately. Furthermore, the incorporation of RAG consistently enhances performance for both vanilla and fine-tuned settings across most model scales, validating the effectiveness of retrieval-augmented strategies in improving model performance.

Notably, for smaller models such as Qwen-VL-2-2B and Qwen-VL-2.5-3B, employing only fine-tuning leads to performance degradation, suggesting that their limited parameter sizes hinder effective learning of domain-specific knowledge through fine-tuning alone. Additionally, for Qwen-VL-2.5-3B model, using RAG without fine-tuning results in a performance drop, likely due to the unique pre-training characteristics of this model. Overall, while fine-tuning or RAG independently can enhance performance in larger-scale models, the best results are consistently achieved by combining these two strategies, underscoring the importance of an integrated approach to maximize VLM effectiveness. From a practical perspective, the combination of fine-tuning and RAG proves particularly suitable for enhancing decision-making capabilities in VLMs. Deploying this optimal configuration can substantially improve VLM performance, with potential applications extending to semantic comprehension, trajectory planning, and other complex autonomous driving tasks.

## 5. Conclusion

In this work, we propose a RAD framework, a novel retrieval-augmented architecture designed to enhance the meta-action decision-making capabilities of VLMs for autonomous driving. Through the integration of fine-tuning VLMs for spatial perception enhancement and BEV image comprehension, RAD effectively enhances VLMs' capability of meta-action decision-making, ensuring higher accuracy, as demonstrated by notable performance gains across key metrics in extensive experimental evaluations.

Moving forward, we aim to extend RAD in three key directions. First, we plan to incorporate more diverse and fine-grained datasets beyond the NuScenes dataset, encompassing more challenging corner cases and real-world scenarios, to further enhance model robustness. Second, we seek to generalize the RAD framework to additional driving tasks, especially trajectory planning and motion control. Third, integrating chain-of-thought and reinforcement learning into the framework will be crucial for improving decision-making depth and adaptability. While fine-tuning and RAG will remain essential for enhancing VLM generalization, these advancements will strengthen the robustness and reliability of autonomous driving systems by leveraging RAG methods to tackle complex real-world tasks.